\documentclass[]{amsart}
\usepackage[colorlinks,urlcolor=blue,linkcolor=blue,citecolor=blue]{hyperref}
\usepackage{graphicx}
\usepackage{amsmath}
\usepackage{amsthm}
\usepackage[foot]{amsaddr}
\usepackage{algorithm}
\usepackage{algorithmic}
\usepackage{caption}
\usepackage{subcaption}
\usepackage{amssymb}
\usepackage[RPvoltages]{circuitikz}
\usepackage{pgfplots,enumerate}
\usepackage{mathtools,bm}
\newcommand{\interp}[1]{\langle #1 \rangle}
\newtheorem{example}{Example}

\newtheorem{cor}{Corollary}
\newtheorem{df}{Definition}
\newtheorem{prp}{Proposition}

\pgfplotsset{compat=1.8}
\newtheorem{theorem}{Theorem}

\tikzset{%
  zeroarrow/.style = {-stealth,dashed},
  onearrow/.style = {-stealth,solid},
  c/.style = {circle,draw,solid,minimum width=2em,
        minimum height=2em},
  r/.style = {rectangle,draw,solid,minimum width=2em,
        minimum height=2em}
}
\usetikzlibrary{positioning}
\begin{document}
\title{Belief Revision in Sentential Decision Diagrams}
\author{Lilith Mattei}
\author{Alessandro Facchini}
\author{Alessandro Antonucci}
\address{Istituto Dalle Molle di Studi sull'Intelligenza Artificiale (IDSIA), Lugano - Switzerland}
\email{\{lilith.mattei,alessandro.facchini,alessandro.antonucci\}@idsia.ch}
\begin{abstract}
Belief revision is the task of modifying a knowledge base when new information becomes available, while also respecting a number of desirable properties. Classical belief revision schemes have been already specialised to \emph{binary decision diagrams} (BDDs), the classical formalism to compactly represent propositional knowledge. These results also apply to \emph{ordered} BDDs (OBDDs), a special class of BDDs, designed to guarantee canonicity. Yet, those revisions cannot be applied to \emph{sentential decision diagrams} (SDDs), a typically more compact but still canonical class of Boolean circuits, which generalizes OBDDs, while not being a subclass of BDDs. Here we fill this gap by deriving a general revision algorithm for SDDs based on a syntactic characterisation of Dalal revision. A specialised  procedure for DNFs is also presented. Preliminary experiments performed with randomly generated knowledge bases show the advantages of directly perform revision within SDD formalism.
\end{abstract}
\maketitle
\section{Introduction}
Belief Revision denotes the operation of partially modifying a knowledge-based system so to consistently incorporate new information that has become available. General well-known standard rationality principles for belief revision have been introduced in \cite{alchourron1985logic}. Several authors have subsequently presented specific revision operators satisfying that framework (e.g., \cite{borgida1984intelligent,dalal1988investigations,satoh1988nonmonotonic}).

When considering a finite propositional language, and thus when a knowledge-based system may be identified with a single formula, an equivalent set of postulates that a revision operator should satisfy to be considered as rational has been outlined in \cite{katsuno1989unified}. However, in such finitary setting, rationality may not be the only desideratum. More specifically, when one is interested in implementing a belief operation in real-world cases, availability of properties such as compactness and tractability of the used representation formalism may become particularly relevant. 

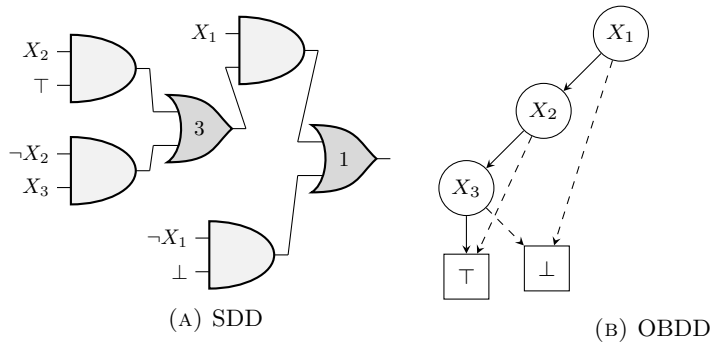
\begin{figure}[htp!]
\centering
\begin{subfigure}[]{0.45\linewidth}
\begin{tikzpicture}[scale=0.8,transform shape]
\draw (0,-2.5) node[and port,fill=black!5] (myand3){};
\draw (0,-4.2) node[and port,fill=black!5] (myand4){};
\draw(1.6,-3.5) node[or port,fill=black!15] (myor2){3};
\draw(2.8,-2.2) node[and port,fill=black!5] (myand1){};
\draw(2.3,-5.6) node[and port,fill=black!5] (myand2){};
\draw(4,-4) node[or port,fill=black!15] (root){1};
\draw (myand1.in 1) node[anchor=east] {$X_1$};
\draw (myand2.in 1) node[anchor=east] {$\neg X_1$};
\draw (myand2.in 2) node[anchor=east] {$\bot$};
\draw (myand3.in 1) node[anchor=east] {$X_2$};
\draw (myand3.in 2) node[anchor=east] {$\top$};
\draw (myand4.in 1) node[anchor=east] {$\neg X_2$};
\draw (myand4.in 2) node[anchor=east] {$X_3$};
\draw (myand1.out) -- (root.in 1) node[]{};
\draw (myand2.out) -- (root.in 2) node[]{};
\draw (myand3.out) -- (myor2.in 1) node[]{};
\draw (myand4.out) -- (myor2.in 2) node[]{};
\draw (myor2.out) -- (myand1.in 2) node[]{};
\end{tikzpicture}
\caption{SDD}
\end{subfigure}
\begin{subfigure}[]{0.45\linewidth}
\begin{tikzpicture}[node distance=0.5cm and 0.5cm]\footnotesize
\node[c] (x1) {$X_1$};
\node[c] (x2) [below left=of x1] {$X_2$};
\node[c] (x3) [below left=of x2] {$X_3$};
\node[r] (1) [below =of x3] {$\top$};
\node[r] (0) [below right=of x3] {$\bot$};
\draw[onearrow] (x1) -- (x2);
\draw[onearrow] (x2) -- (x3);
\draw[onearrow] (x3) -- (1);
\draw[zeroarrow] (x1) -- (0);
\draw[zeroarrow] (x2) -- (1);
\draw[zeroarrow] (x3) -- (0);
\end{tikzpicture}
\caption{OBDD}
\end{subfigure}
\caption{A SDD and an OBDD representing the formula $X_1 \wedge (\neg X_2 \vee X_3)$. The SDD is just a Boolean circuit alternating disjunctive and conjunctive gates with literals and constants as inputs. The BDD is a decision diagram whose inner nodes (circles) are decision nodes associated with the true (solid arcs) and false (dashed arcs) states of the corresponding variable and with constants on the leaves.}
\label{fig:sdd-obdd}
\end{figure}

Originally presented in \cite{darwiche2011sdd}, \emph{Sentential Decision Diagrams} (SDDs) are a powerful yet compact and empirically efficient representation formalism for propositional knowledge bases. They can be regarded as a generalization of \emph{Ordered Binary Decision Diagram} (OBDDS) 
\cite{bryant1992symbolic}, sharing the canonicity property and the fact that Boolean combinations take only polynomial time. Fast compilation schemes have been derived for SDDs \cite{oztok2015top}, making the resulting models typically smaller than the corresponding OBDDs \cite{Bova_2016}. Examples of these formalisms are in Figure \ref{fig:sdd-obdd}.

As a matter of fact, the analysis of belief revision on specific representation formalisms received little attention. To the best of our knowledge the only study of this type has been performed by Gorogiannis and Ryan in \cite{gorogiannis2002implementation} for so called \emph{Binary Decision Diagrams} (BDDs), a very general class of models encompassing OBDDs, but not SDDs.

In this work we perform an analogous analysis with respect to SDDs, when the specific belief operator corresponds to the one introduced by Dalal in \cite{dalal1988investigations}. There, Dalal defines a semantic measure for minimal change, whose idea, roughly, is to change the truth assignment to the smallest number of propositional variables, and introduces a syntactic revision operator that behaves accordingly. The technique proposed by Dalal has however several drawbacks. In particular, it requires the performance of multiple satisfiability checks, a notorious NP-complete task. 

In this paper,  we partially overcome such difficulty by adapting Dalal revision to the particular representation language under consideration. In doing so, we show that the computation of \emph{resolvents}, the units on which the syntactic revision operator is built, is polytime in SDDs. The same holds for the satisfiability check, because of the linear complexity of model counting in SDDs. Finally, we deliver two algorithms implementing Dalal revision: the first, for the general case, has exponential worst-case complexity, whereas the second considers the particular case in which the new information is a DNF and runs slightly faster.

The paper is organized as follows. We first outline the basic concepts of belief revision and present Dalal's approach. Compared to the classical formulation, we focus on the syntactic formulation and derive some new representation results for higher order relaxations. Then, after providing the necessary background on SDDs, our contribution is presented and it corresponds to implement Dalal revision in SDDs by exploiting the aforementioned syntactic formulation. Finally, we draw preliminary experiments performed with randomly generated knowledge bases, in order to show the computational advantages of performing revision within the SDD framework.

\section{Background}
\subsection{Basic Terminology and Notation.}
\subsubsection{Languages and Interpretations.} We fix a propositional language $\mathcal{L}$ over a finite set of Boolean variables $\bm{X}$ of cardinality $n$. A $\mathcal{L}$-interpretation $w$ is a truth assignment for every variable in $\bm{X}$. We describe a $\mathcal{L}$-interpretation $w$ by listing the variables taking true value according to $w$. The interpretation of a $\mathcal{L}$-formula is obtained inductively by the standard truth tables. The distance $\Delta(w, w')$ between two $\mathcal{L}$-interpretations $w, w'$ is the number of variables in which they differ, i.e.,
\begin{align*}
\Delta(w,w') := |\left\{X \in \bm{X} \left| w(X) \neq w'(X) \right.\right\}|\,.
\end{align*}

We assume that the language contains the Boolean constants always taking the false or true values and denoted, respectively, as $\bot$ and $\top$. For a formula $\psi\in\mathcal{L}$, we denote as $mod(\psi)$ the set of its models, that is, the set of $\mathcal{L}$-interpretations making $\psi$ true. Formulae $\phi,\psi\in\mathcal{L}$ are logically equivalent, written $\phi \equiv \psi$, whenever $mod(\phi)=mod(\psi)$. As usual, entailment $\phi \models \psi$ means that $mod(\phi) \subseteq mod(\psi)$. In the following, by $\psi^-_X$, resp. $\psi^+_X$, we denote the formula obtained by substituting any occurrence of $X$ in $\psi$ with constant $\bot$, resp. with $\top$.

\subsubsection{Knowledge Bases.} A \textit{knowledge base} (KB) is a  set of formulae in a propositional language $\mathcal{L}$, i.e., a $\mathcal{L}$-theory describing the possible states of the world via its models. In what follows we always assume a KB to be finite, and therefore it is identified with the formula, denoted by $\psi$, obtained by taking the conjunction of all its members. 

\subsection{Coping with New Information}

\subsubsection{Belief Revision.} Consider a KB represented by the formula $\psi$. The task to modify the current KB $\psi$ in the light of a new piece of information $\mu$ goes under the name of \textit{belief change}. Following \cite{alchourron1985logic}, we distinguish different kinds of belief changes. \textit{Expansion} consists in adding $\mu$ to $\psi$ and taking its semantic closure, disregarding whether the obtained KB is consistent or not. In \emph{contraction} one simply eliminates from $\psi$ the sentences inconsistent with $\mu$. The belief change we are interested in is called \textit{belief revision}: in this case $\mu$ is inconsistent with $\psi$, so in order to add it to the latter and obtain a new, consistent, KB, we have to get rid of the inconsistencies. Let us denote by $\psi \circ \mu$ the revised knowledge obtained by revising the old knowledge $\psi$ in the light of the new information $\mu$. In practice a belief revision algorithm receives $\psi$ and $\mu$ in input and returns the revised knowledge $\psi \circ \mu$ as output such that the latter  is obtained by minimally changing the models of $\psi$ in such a way that $\mu$ holds in at least some of them. A number of desirable principles for such algorithms is reported here below. 

\subsubsection{Rationality Postulates.} As we are dealing with finite propositional languages, we consider the postulates of \cite{katsuno1989unified}, characterising rationality of a belief revision operator.
\begin{description}
\item[(R1)] $\psi \circ \mu$ implies $\mu$.
\item[(R2)] If $\psi \land \mu$ is satisfiable, then $\psi \circ \mu \equiv \psi \land \mu$.
\item[(R3)] If $\mu$ is satisfiable, then $\psi \circ \mu$ is also satisfiable.
\item[(R4)] If both $\psi \equiv \psi'$ and $ \mu \equiv \mu'$, then  $\psi \circ \mu \equiv \psi' \circ \mu'$.
\item[(R5)] $(\psi\circ \mu) \land \phi$ implies $\psi\circ (\mu \land \phi)$.
\item[(R6)] If $(\psi\circ \mu) \land \phi$ is satisfiable, then $\psi\circ (\mu \land \phi)$ implies $(\psi\circ \mu) \land \phi$.
\end{description}

\subsubsection{Dalal Revision.}
Here we describe revision as presented by Dalal in \cite{dalal1988investigations}. Revision is initially defined in terms of models of the KB, i.e., at the pure semantic level. In a second moment, an equivalent syntactic characterisation is also provided.

\begin{df}\label{def::spheres}
Let $w$ be a $\mathcal{L}$-interpretation and $\mathcal{A}$ a set of $\mathcal{L}$-interpretations. For each integer $\ell \geq 0$, denote by $g^\ell(w)$ the set of $\mathcal{L}$-interpretations at distance at most $\ell$ from $w$, i.e.,
\begin{align*}
g^\ell(w) := \left\{ w' \left| \Delta(w,w') \leq \ell \right.\right\}
\end{align*}
and by $g^\ell(\mathcal{A})$ the set of $\mathcal{L}$-interpretations at distance at most $\ell$ from at least one element of $\mathcal{A}$, i.e., 
\begin{align*}
g^\ell(\mathcal{A}) := \mathop{\cup}_{w\in \mathcal{A}}g^\ell(w)\,.
\end{align*}
\end{df}

We are ready to define a \emph{relaxation operator} $G^\ell$, acting on a KB $\psi$. 

\begin{df}\label{def::relaxationop}
Consider a KB $\psi$ and an integer $i\geq 0$. The $\ell^{\mathrm{th}}$ relaxation of $\psi$, denoted by $G^\ell(\psi)$, is the $\mathcal{L}$-formula that satisfies the condition:
\begin{align*}
mod(G^\ell(\psi)) = g^\ell(mod(\psi))\,.
\end{align*}
\end{df}

The models of $G^\ell(\psi)$ are all interpretations differing in at most $\ell$ variables from (at least) one model of $\psi$. The models of $\psi$ are clearly included in this collection, justifying the name of the operator. 

The above defined operators admit a simple recursive formulation as shown by the following result.

\begin{prp}\label{prop::recursivity}
Consider a set of interpretations $\mathcal{A}$, a KB $\psi$ and an integer $\ell\geq 0$. Then:
\begin{align*}
g^\ell(\mathcal{A}) = g^1(g^{\ell-1}(\mathcal{A}))\,,
\end{align*}
and
\begin{align*}
 G^\ell(\psi) = G^1(G^{\ell-1}(\psi))\,.
\end{align*}
\end{prp}

We can now get back to our initial purpose. Given a KB $\psi$ and new information $\mu$, to give a definition of $\psi \circ \mu$ respecting the desiderata in the previous section. In order to satisfy R2, if $\mu$ is consistent with $\psi$, we shall simply take $\psi \circ \mu = \psi \wedge \mu$. Accordingly, in the definition below, we assume $\mu$ to be inconsistent with $\psi$. 

\begin{df}\label{def::revision}
For KB $\psi$ and new information $\mu$ inconsistent with $\psi$,
\begin{align*}
\psi \circ \mu := G^k(\psi) \wedge \mu\,,
\end{align*}
where $k$ is the smallest integer $\ell \in \{1,\dots, n\}$ such that $g^\ell(mod(\psi))$ contains a model of $\mu$.
\end{df}

Dalal revision consists in (possibly) repeated relaxations of the initial KB $\psi$: if $G^1(\psi)$ is consistent with the new information $\mu$, then $\psi \circ \mu = G^1(\psi) \wedge \mu$, otherwise we consider $G^2(\psi)=G^1(G^1(\psi))$ and so on. Notice that when $k=n$, we obtain $\psi \circ \mu = \mu$.
We call $k$ the \emph{order} of a Dalal revision. 

The following is a classical example of Dalal revision.

\begin{example}
Assume $\bm{X} = \{X,Y\}$, a KB represented by $\psi= X\wedge \neg Y$ and new information $\mu = Y$. The new information is inconsistent with the KB. We start by considering $mod(G^1(\psi))$. In order to describe the latter, we need $g^1_{\bm{X}}(mod(\psi))$. Since $mod(\psi) = \{ \{X\}\}$, the models differing from those of $\psi$ on at most one variable are $\{X\}, \{X,Y\}$ and $\emptyset$. Notice that $\{X,Y\}$ is also a model of $\mu$, thus $\psi \circ \mu \equiv X\wedge Y$.
\end{example}

It is a simple exercise to prove the following result.
\begin{prp}
Dalal revision satisfies the postulates R1-R6.
\end{prp}

\subsubsection{Syntactic Characterisation.} In the previous section we have seen how to possibly revise the KB in a semantic way, i.e., the revised KB is defined in terms of its models. In \cite{dalal1988investigations}, however, Dalal also provides a syntactic characterisation of the derivation of $G^1(\psi)$ from $\psi$. We present here below the various steps for generalizing such characterisation to $G^k(\psi)$. In this way the revision process defined by Definition \ref{def::revision} will mainly reduce to a syntactic manipulation of the initial formula.

The following observation is straightforward.
\begin{prp}\label{lemma::res}
Let  $\psi$ be an  $\mathcal{L}$-formula and $X \in \bm{X}$. 
Then $\psi^-_X$ and  $\psi^+_X$ do not contain variable $X$ and
$$\psi \equiv (\neg X \wedge \psi^-_X) \vee (X \wedge \psi^+_X)\,.$$
\end{prp}

Based on this, Dalal introduces the following definition.
\begin{df}
Let $\psi$ be a $\mathcal{L}$-formula, $X \in \bm{X}$. Then:
\begin{align*}
res_{X}(\psi) := \psi^-_X \vee \psi^+_X\,,
\end{align*}
is called the \emph{resolvent} of $\psi$ with respect to $X$.
\end{df}
In what follows, we call $\psi^-_X$ and $\psi^+_X$ the \emph{semi-resolvents} of $\psi$ with respect to $X$. 

The next result provides us with a more explicit formulation for $G^1(\psi)$ based on the previous definitions. 

\begin{prp}[\cite{dalal1988investigations}]\label{thm::dalal}
Let $\psi$ be a $\mathcal{L}$-formula over variables $\bm{X}$. Then:
\begin{align*}
G^1(\psi) & \equiv \mathop{\bigvee}_{X\in \bm{X}} res_{X}(\psi) \\
& \equiv \bigvee_{X\in \bm{X}}
\bigvee_ {\sigma \in \{+,-\} } \psi^\sigma_{X}\,.
\end{align*}
\end{prp}

For the purposes of this paper, we want to generalize this classical result to higher revision orders. This might be achieved by the notion of \emph{higher order semi-resolvents} introduced by the following definitions. 

\begin{df}\label{defdisj}
Let  $\psi$ be an  $\mathcal{L}$-formula over variables $\bm{X}$, $X_1, \dots, X_\ell \in \bm{X}$. Then: 
\begin{align*}
res_{X_1, \dots, X_\ell}(\psi) := \mathop{\bigvee}_{\bm{\sigma}\in \{+,-\}^\ell} \psi_{X_1,\dots,X_\ell}^{\bm{\sigma}}\,,
\end{align*}
is called the \emph{resolvent of order} $\ell$, or $\ell$-\emph{resolvent} of $\psi$ with respect to $X_1,\dots,X_\ell$, whereas the formulas  $\psi_{X_1,\dots,X_\ell}^{\bm{\sigma}}$
are called the \emph{semi-resolvents of order} $\ell$, or $\ell$-\emph{semi-resolvents} of $\psi$ with respect to $X_1,\dots,X_\ell$.
\end{df}

First of all notice that Lemma \ref{lemma::res} can be straightforwardly generalised as follows. 
\begin{prp}\label{lemma::resk}
Let  $\psi$ be an  $\mathcal{L}$-formula over variables $\bm{X}$. Given $X_1, \dots, X_\ell \in \bm{X}$, it holds that   $\psi_{X_1,\dots,X_\ell}^{\bm{\sigma}}$ does not contain variables $X_1,\dots,X_\ell$. Moreover
$$\psi \equiv 
\bigvee_{X_1,\dots,X_\ell \in \bm{X}} \bigvee_{\bm{\sigma}\in \{+,-\}^\ell}
\left(\psi_{X_1,\dots,X_\ell}^{\bm{\sigma}} \wedge \bigwedge^\ell_{i=1}X_i^{\bm{\sigma}(i)}\right)$$
where $ X_i^{\bm{\sigma}(i)}=X_i$ if $\bm{\sigma}(i)=+$, and $ X_i^{\bm{\sigma}(i)}=\neg X_i$ otherwise.
\end{prp}


The next proposition follows directly from the definition of the $\ell$-resolvent of a formula in terms of its $\ell$-semi-resolvents.

\begin{prp}\label{prop::properties}
Let $\psi$ be a $\mathcal{L}$-formula over variables $\bm{X}$, and $X, X_1,\dots X_\ell \in \bm{X}$. The following properties hold:
\begin{align*}
 res_{X}(res_{X}(\psi)) \equiv res_{X}(\psi)   
\end{align*}
and
 \begin{align*}
     res_{X_1}(\dots (res_{X_\ell}(\psi))\dots )   \\
\equiv  res_{\pi(X_1)}(\dots (res_{\pi(X_\ell)}(\psi))\dots ) \\
\equiv \mathop{\bigvee}_{\bm{\sigma \in \{+,-\}^\ell}} \psi_{X_1,\dots,X_\ell}^{\bm{\sigma}}, 
\end{align*}
where $\pi(X_1),\dots \pi(X_\ell)$ is an arbitrary permutation of $X_1,\dots X_\ell$.
\end{prp}

Notice that, by Definition \ref{defdisj}, the last property of  Proposition \ref{prop::properties} implies that:
\begin{equation}\label{eq::resk}
res_{X_1, \dots, X_\ell}(\psi) \equiv res_{X_1}(\dots (res_{X_\ell}(\psi))\dots )\,.
\end{equation}
Moreover, $G^\ell$ distributes over disjunctions as shown by the following result.

\begin{prp}\label{prop::distributivity}
Given formulae $\psi_1,\dots,\psi_m$, for each $\ell\geq 0$:
\begin{align*}
G^\ell\left(\mathop{\bigvee}_{j=1}^m \psi_j\right) \equiv  \mathop{\bigvee}_{j=1}^m G^\ell(\psi_j)\,.
\end{align*}
\end{prp}

Propositions \ref{thm::dalal}, \ref{prop::properties} and  \ref{prop::distributivity}, allow us to state the following corollary. It generalises Proposition \ref{thm::dalal} in terms of higher order semi-resolvents.

 
 
 \begin{cor}\label{cor::Gksemires} Let $\ell \in \{1, \dots, n\}$, then
\begin{align*}
G^\ell(\psi) 
& \equiv \mathop{\bigvee}_{X_1,\dots,X_\ell \in \bm{X}}res_{X_1}(\dots (res_{X_\ell}  (\psi))\dots)
\\
& \equiv \bigvee_{X_1,\dots,X_\ell \in \bm{X}}\bigvee_{\bm{\sigma}\in \{+,-\}^\ell } \psi_{X_1,\dots,X_\ell}^{\bm{\sigma}}.
\end{align*}
\end{cor}
 
Notice that the previous result is telling us that
$G^\ell(\psi)$ is tantamount to the disjunction of $\frac{n!}{\ell! (n-\ell)! }\cdot 2^\ell$ semi-resolvents of order $\ell$.

Corollary \ref{cor::Gksemires} is at the base of our strategy for applying Dalal revision to SDDs, as we will see in the dedicated section.



\section{Sentential Decision Diagrams}
In this section we recall the basic definitions and characterisations for \emph{sentential decision diagrams} (SDDs). The general revision procedure described in the previous section will be specialized to SDDs in the next section. To achieve that, we first need to generalise the notion of order over a set of variables by means of the following definition.

\begin{df}[Vtree]
A \emph{vtree} for a finite set of Boolean variables $\bm{X}$ is a full binary tree whose leaves are in one-to-one correspondence with the elements of $\bm{X}$. 
\end{df}

Given a vtree internal node $v$, denote by  $v^l$ (resp., $v^r$) its left (right) child. In what follows we will call a vtree by its root node label, so that for vtree $v$, $v^l$ (resp., $v^r$) denotes the vtree rooted at the left (resp., right) child of $v$.

\begin{figure}[htp!]
\centering
\begin{subfigure}[b]{3cm}
\begin{tikzpicture}[scale=0.8,transform shape,level distance=1cm,level 1/.style={sibling distance=1.5cm},level 2/.style={sibling distance=1cm}]
\node {$3$}child {node {$1$} child {node {$L$}}
child {node {$K$}}} child {node {$5$}
child {node {$P$}} child {node {$A$}}};
\end{tikzpicture}
\caption{\label{fig:vtree1}}
\end{subfigure}
\quad\quad
\begin{subfigure}[b]{3cm}
\begin{tikzpicture}[scale=0.8,transform shape,level distance=1cm,
level 1/.style={sibling distance=1cm},level 2/.style={sibling distance=1cm}]
\node {$3$} child {node {$1$} child {node {$L$}} child { node {$5$} child {node {$K$}} child {node {$P$}}}} child {node {$A$}};
\end{tikzpicture}
\caption{\label{fig:vtree2}}
\end{subfigure}
\caption{Two vtrees over $\bm{X}=\{L,K,P,A\}$.}\label{fig:vtree}
\end{figure}
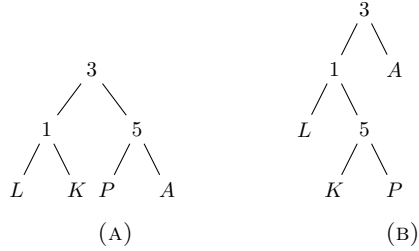

In-order vtree traversal induces a total order on its variables, but two distinct vtrees might lead to the same order. This is for instance the case for the two vtrees in Figure \ref{fig:vtree}.

Based on the notion of vtree, we provide the following
definition of sentential decision diagram \cite{darwiche2011sdd}.

\begin{df}[SDD]\label{def:sdd}
A \emph{sentential decision diagram} $S$ \emph{normalised for vtree} $v$ and the corresponding Boolean formula $\interp{S}$ are inductively defined as follows.
\begin{itemize}
\item  If $v$ is a leaf, let $X$ be the variable attached to $v$; then $S$ is either a \emph{constant}, i.e., $S \in \{ \bot,  \top \}$, with $\interp{\bot} = \bot$ and $\interp{\top} = \top$,  or a \emph{literal}, i.e., $S \in \{X, \neg X \}$, with 
$\interp{X}:= X$ and $\interp{\neg X}:=\neg X$.

\item If $v$ is not a leaf, then  $S := \{ (p_i,s_i)\}_{i=1}^k$, where the  \emph{primes} $\{p_i\}_{i=1}^\ell$ and the \emph{subs} $\{s_i\}_{i=1}^k$ are SDDs normalised for $v^l$ and $v^r$, respectively, such that $\{\interp{p_i}\}_{i=1}^k$ is a partition; in this case $\interp{S}:=\mathop{\bigvee}_{i=1}^k \interp{p_i} \wedge \interp{s_i}$.
\end{itemize}

\end{df}

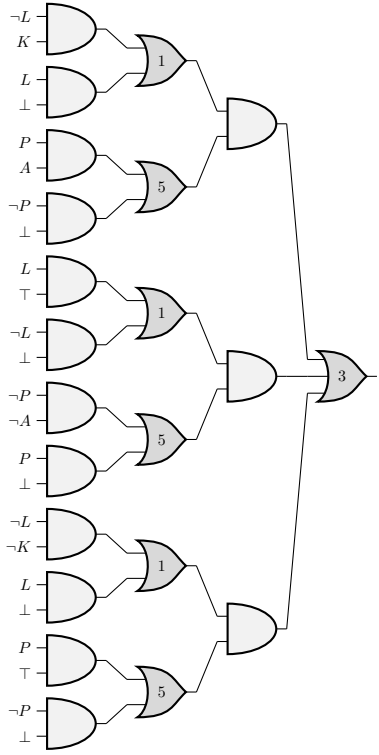
\begin{figure}[htp!]
\centering
\begin{tikzpicture}[scale=0.6,transform shape]
\draw (0,0) node[and port,fill=black!5] (myand1){};
\draw (0,-1.4) node[and port,fill=black!5] (myand2){};
\draw (0,-2.8) node[and port,fill=black!5] (myand3){};
\draw (0,-4.2) node[and port,fill=black!5] (myand4){};
\draw (0,-5.6) node[and port,fill=black!5] (myand5){};
\draw (0,-7.0) node[and port,fill=black!5] (myand6){};
\draw (0,-8.4) node[and port,fill=black!5] (myand7){};
\draw (0,-9.8) node[and port,fill=black!5] (myand8){};
\draw (0,-11.2) node[and port,fill=black!5] (myand9){};
\draw (0,-12.6) node[and port,fill=black!5] (myand10){};
\draw (0,-14) node[and port,fill=black!5] (myand11){};
\draw (0,-15.4) node[and port,fill=black!5] (myand12){};
\draw(2,-0.7) node[or port,fill=black!15] (myor1){1};
\draw(2,-3.5) node[or port,fill=black!15] (myor2){5};
\draw(2,-6.3) node[or port,fill=black!15] (myor3){1};
\draw(2,-9.1) node[or port,fill=black!15] (myor4){5};
\draw(2,-11.9) node[or port,fill=black!15] (myor5){1};
\draw(2,-14.7) node[or port,fill=black!15] (myor6){5};
\draw(4,-2.1) node[and port,fill=black!5] (myand1b){};
\draw(4,-7.7) node[and port,fill=black!5] (myand2b){};
\draw(4,-13.3) node[and port,fill=black!5] (myand3b){};
s\draw(6,-7.7) node[or port,fill=black!15,number inputs=3] (root){3};
\draw (myand1.in 1) node[anchor=east] {$\neg L$};
\draw (myand1.in 2) node[anchor=east] {$K$};
\draw (myand2.in 1) node[anchor=east] {$L$};
\draw (myand2.in 2) node[anchor=east] {$\bot$};
\draw (myand3.in 1) node[anchor=east] {$P$};
\draw (myand3.in 2) node[anchor=east] {$A$};
\draw (myand4.in 1) node[anchor=east] {$\neg P$};
\draw (myand4.in 2) node[anchor=east] {$\bot$};
\draw (myand5.in 1) node[anchor=east] {$L$};
\draw (myand5.in 2) node[anchor=east] {$\top$};
\draw (myand6.in 1) node[anchor=east] {$\neg L$};
\draw (myand6.in 2) node[anchor=east] {$\bot$};
\draw (myand7.in 1) node[anchor=east] {$\neg P$};
\draw (myand7.in 2) node[anchor=east] {$\neg A$};
\draw (myand8.in 1) node[anchor=east] {$P$};
\draw (myand8.in 2) node[anchor=east] {$\bot$};
\draw (myand9.in 1) node[anchor=east] {$\neg L$};
\draw (myand9.in 2) node[anchor=east] {$\neg K$};
\draw (myand10.in 1) node[anchor=east] {$L$};
\draw (myand10.in 2) node[anchor=east] {$\bot$};
\draw (myand11.in 1) node[anchor=east] {$P$};
\draw (myand11.in 2) node[anchor=east] {$\top$};
\draw (myand12.in 1) node[anchor=east] {$\neg P$};
\draw (myand12.in 2) node[anchor=east] {$\bot$};
\draw (myand1.out) -- (myor1.in 1) node[]{};
\draw (myand2.out) -- (myor1.in 2) node[]{};
\draw (myand3.out) -- (myor2.in 1) node[]{};
\draw (myand4.out) -- (myor2.in 2) node[]{};
\draw (myand5.out) -- (myor3.in 1) node[]{};
\draw (myand6.out) -- (myor3.in 2) node[]{};
\draw (myand7.out) -- (myor4.in 1) node[]{};
\draw (myand8.out) -- (myor4.in 2) node[]{};
\draw (myand9.out) -- (myor5.in 1) node[]{};
\draw (myand10.out) -- (myor5.in 2) node[]{};
\draw (myand11.out) -- (myor6.in 1) node[]{};
\draw (myand12.out) -- (myor6.in 2) node[]{};
\draw (myor1.out) -- (myand1b.in 1) node[]{};
\draw (myor2.out) -- (myand1b.in 2) node[]{};
\draw (myor3.out) -- (myand2b.in 1) node[]{};
\draw (myor4.out) -- (myand2b.in 2) node[]{};
\draw (myor5.out) -- (myand3b.in 1) node[]{};
\draw (myor6.out) -- (myand3b.in 2) node[]{};
\draw (myand1b.out) -- (root.in 1) node[]{};
\draw (myand2b.out) -- (root.in 2) node[]{};
\draw (myand3b.out) -- (root.in 3) node[]{};
\end{tikzpicture}
\caption{A SDD for the KB in Equation \eqref{eq:kbkisa} normalized for the vtree in Figure~\ref{fig:vtree1}.}
\label{fig:toy}
\end{figure}

SDDs have a recursive structure: each prime and sub in a SDD $S$ normalised for vtree $v$ is in turn a SDD - normalised for a sub-vtree of $v$ - and we call it a \emph{node} of $S$. A node $n$ of $S$ can be either a \emph{terminal} node, when it is normalised for some $v$'s leaf, or a \emph{decision} node otherwise. In a decision node $\{(p_i,s_i)\}_{i=1}^k$,  each pair $(p_i, s_i)$ is an \emph{element} of the node and $k$ is its \emph{size}. The \emph{size} of a SDD $S$, denoted $|S|$, is the sum of the sizes of its decision nodes. At the interpretation level, each decision node represents a (exclusive, as the primes form a partition) disjunction, while each element is a conjunction between a prime and the corresponding sub. We can therefore intend a SDD $S$ as a rooted logic circuit, providing a representation of the formula $\langle S \rangle$. Figure \ref{fig:toy} depicts an example of such circuit representation. Labels on the decision nodes denote the vtree nodes for which the sub-SDD is normalized. 

\begin{example}\label{ex:kisa}
As a demonstrative example to be used in this work to illustrate our belief revision procedure, we consider the simple SDD over four variables introduced by
\cite{kisa2014probabilistic}. Its variables are the subjects related to the enrolling of a first-year student, namely \emph{logic} ($L$), \emph{knowledge representation} ($K$), \emph{probability} ($P$) and \emph{AI} ($A$). Faculty rules give constraints on the study plan corresponding to the KB: 
\begin{equation}\label{eq:kbkisa}
\psi := (L \vee P) \wedge (A \Rightarrow P) \wedge (K \Rightarrow  A \vee L )\,.
\end{equation}
Figure \ref{fig:toy} depicts a circuit representation of a SDD $S$ normalized for the vtree in Figure \ref{fig:vtree1} and such that $\langle S \rangle \equiv \psi$.
\end{example}

Once a propositional KB $\psi$ has been compiled as a SDD $S$, inference schemes to check satisfiability of a formula and solve model counting can be implemented in linear time with respect to the size $|S|$ of $S$ by a simple bottom-up traversal of the circuit \cite{vdb2015role}. Accordingly, in the pseudocode of the algorithms presented in the next section, we will denote as ${\tt satisfies}(S,c)$ the Boolean function returning true if and only if $c \wedge \psi$ is satisfiable where $c$ is a conjunction of literals, while the model count operator ${\tt mc}(S)$ gives the number of models satisfying $\psi$, where $\psi \equiv \langle S\rangle$. 

Exactly as OBDDs, SDDs can be combined by any binary Boolean operator provided that both SDDs are normalized for the same vtree. Notation ${\tt apply}(S_1,S_2,\mathrm{op})$ refers to the procedure returning a SDD $S$ such that $\langle S \rangle \equiv \mathrm{op}(\langle S_1 \rangle, \langle S_2 \rangle)$, where $\mathrm{op}$ is any binary Boolean operator. Note that both the size of the output and the running time are $|S|=O(|S_1|\cdot |S_2|)$. An iterative application of the ${\tt apply}$ function allows for a trivial bottom-up compilation of any propositional formula \cite{vdb2015role}. Top-down compiler have been also proposed and proved to achieve faster compilation times \cite{oztok2015top}. A dynamic search over the vtree space can be also considered to find the ones leading to the smaller models \cite{choi2013dynamic}.

We instead denote as ${\tt expand}(S)$ the procedure replacing all the decision nodes such that the primes are sub-SDDs over a single variable $X$ and made of a single element having $\top$ as unique prime, with two elements having $X$ and $\neg X$ as primes and the original sub as new subs. Vice versa ${\tt compress}$ denotes the operation of merging together elements with the same sub by taking the union of the primes. Finally, $S':=S.{\tt replace}(s,s')$ is just a SDD obtained from $S$ by replacing its sub-SDD $s$ with a new SDD $s'$. It is a trivial remark that if both $s$ and $s'$ are normalized for the same (sub)vtree, $S$ and $S'$ are also normalized for the same vtree.

\section{Dalal Revision in SDDs}

In this section we implement Dalal revision in SDDs. We consider both the original KB $\psi$ and a new piece of information $\mu$ to be encoded in two SDDs denoted respectively as $S$, and $S'$. Unlike \cite{gorogiannis2002implementation}, where revision in BDDs is performed by a purely semantic approach, we perform Dalal revision in SDDs by exploiting the syntactic characterisation of the relaxation operator described in Corollary \ref{cor::Gksemires}.

\subsection{Semi-resolvents Computation.} 
In order to achieve that, we start by building SDDs representing the semi-resolvents of $\psi$. For each $X\in \bm{X}$, Algorithm \ref{alg:ReSDD} performs local transformations on $S$ to obtain SDD $S^+_X$ representing the semi-resolvent of $\psi^+_X$. The procedure to build $S^-_X$ such that $\langle S^-_X \rangle \equiv \psi^-_X$ is analogous.

Subroutine ${\tt parentof}$ finds the parent of the leaf of a vtree associated with a variable, while ${\tt decisions}$ returns all the decision nodes normalised for a given sub-vtree. Both these procedures can be implemented in linear time with respect to $|S|$ by a DFS traversal. As the operations in lines 7, 12 and 14 might make a prime inconsistent, we prune the branches of this kind. Finally, it is easy to see that
the size of the output is smaller or equal to the one of the input. 


\begin{algorithm}[htp!]\caption{\texttt{SReSDD}${}^{+}$ (Semi-Resolvents Computation in SDDs)}
\label{alg:ReSDD}
\textbf{Input}: $X\in\bm{X}$, SDD $S$ on $\bm{X}$ norm. for $v$ s.t. $\langle S \rangle\equiv\psi$.\\
\textbf{Output}: SDD $S^+_X$ norm. for $v$, s.t. $\langle S^+_X\rangle\equiv \psi^+_X$.\\
\begin{algorithmic}[1]
\STATE $S^+_X \leftarrow {\tt expand}(S)$
\STATE $w\leftarrow {\tt parentof}(X,v)$
\STATE $(i_1,\ldots,i_M) \leftarrow {\tt decisions}(S,w)$ 
\FOR{$m \in 1,\ldots,M$}
\IF{$X \in w^l$}
\STATE {\color{black!50}{// $i_m= \{(X,s_m^+),(\neg X,s_m^-)\}$}} 
\STATE $S^+_X \leftarrow S^+_X.{\tt replace}(s_m^-,s_m^+)$
\STATE ${\tt compress}(i_m)$
\ELSE
\STATE {\color{black!50}{// $i_m= \{ (p_m^t,s_m^t)\}_{t=1}^{L(m)}$}}
\FOR{$t \in 1:L(m)$}
\IF{$s_m^t = X$}
\STATE $S^+_X \leftarrow S^+_X.{\tt replace}(s_m^t, \top)$
\ELSIF{$s_m^t \leftarrow \neg X$}
\STATE $S^+_X \leftarrow S^+_X.{\tt replace}(s_m^t, \bot)$
\ENDIF
\ENDFOR
\STATE ${\tt compress}(i_m)$
\ENDIF
\ENDFOR
\STATE \textbf{return} $S^+_X$ 
\end{algorithmic}
\end{algorithm}


The following result guarantees that the algorithm properly returns a SDD representing the semi-resolvent. 

\begin{theorem}\label{thm::resdd}
Let $\psi$ be a $\mathcal{L}$-formula, $\bm{X}$ be the variables occurring in $\psi$, $v$ a vtree for $\bm{X}$, and $S$ a SDD normalized for $v$ representing $\psi$ (i.e., $\langle S\rangle \equiv\psi$). For each $X\in\bm{X}$, the SDD $S^+_X$ returned by Algorithm \ref{alg:ReSDD} is normalized for $v$ and represents $\psi^+_X$. 
\end{theorem}

The following result gives a bound to the complexity of Algorithm \ref{alg:ReSDD}. 

\begin{prp}\label{prop::complexityresdd}
\texttt{ReSDD} runs in time $O(|S|)$ on input SDD $S$.


\end{prp}

\subsection{Higher Order Semiresolvents.} 
From Definition \ref{defdisj}, it is immediate to notice that a SDD representing a higher order semi-resolvent can be obtained by iterated application of Algorithm \ref{alg:ReSDD} to the initial SDD $S$. Given $X_1,\dots,X_\ell \in \bm{X}$, $\bm{\sigma}\in \{+,-\}^\ell$, an SDD $S^{\bm{\sigma}}_{X_1,\dots,X_\ell}$ representing $\psi^{\bm{\sigma}}_{X_1,\dots,X_\ell}$ is obtained as:
\begin{multline*}
S^{\bm{\sigma}}_{X_1,\dots,X_\ell} =
\texttt{SReSDD}^{\bm{\sigma}(1)}(\dots\\
\texttt{SReSDD}^{\bm{\sigma}(\ell-1)}(\texttt{SReSDD}^{\bm{\sigma}(\ell)}(S,X_\ell), X_{\ell-1}, )\\
\dots)\,.
\end{multline*}
As an obvious consequence of the results in the previous section, the size $|S^{\bm{\sigma}}_{X_1,\dots,X_\ell}|$
of the SDD of a $\ell$-semi-resolvent is still smaller or equal to $|S|$. The same holds for the computation time, which remains linear in the size of $|S|$.

\subsection{Revision Algorithm.}
The discussion in the previous section leads to a syntactic approach to Dalal revision in SDDs. We first compute all the semi-resolvents and check their compatibility with $\mu$. If no compatible semi-resolvent exists, we perform an analogous check on the semi-resolvents of order two and so on. Once the necessary level of relaxation $\ell=k$ is detected, the SDD corresponding to the disjunction of the compatible $k$-semi-resolvents is computed, and the SDD of its conjunction with $\mu$ is finally returned in output. The procedure is depicted by Algorithm \ref{alg:BRevSDD}.
Disjunctions (line 8) and conjunctions (lines 7 and 17) at the SDD level are implemented by the $\tt apply$ operator for SDDs. Model counting (line 7) is used to check compatibility. The two nested loops (starting in lines 5 and 6) enumerate all the distinct $\ell$-semi-resolvents as in Corollary \ref{cor::Gksemires}. Their computation is achieved by Algorithm \ref{alg:ReSDD} and a caching scheme can be considered to obtain the semi-resolvents of order $\ell$  from those of order $\ell-1$.



The following result guarantees that the algorithm properly implements Dalal revision in SDDs.

\begin{algorithm}[htb!]
\caption{\texttt{BRevSDD} (Belief Revision in SDDs)}
\label{alg:BRevSDD}
\textbf{Input}: SDDs $S$ and $S'$ on $\bm{X}$ (with $var(\psi)=\bm{X}$) normalized for $v$ and such that $\langle S \rangle = \psi$ and $\langle S' \rangle = \mu$.\\
\textbf{Output}: SDD $S''$ on $\bm{X}$ norm. for $v$ s.t. $\langle S'' \rangle = \psi \circ \mu$.\\
\begin{algorithmic}[1] 
\STATE revised $\leftarrow$ false
\STATE $\ell \leftarrow 1$
\STATE $S'' \leftarrow {\tt compile}(\bot,v)$
\WHILE{{\bf not} revised}
 \FOR{$\{X_1,\dots,X_\ell\} \subseteq \bm{X} $}
    \FOR{$\bm{\sigma} \in \{+,-\}^\ell$}
        \IF{${\tt mc}({\tt apply}(S^{\bm{\sigma}}_{X_1,\dots,X_\ell},S',\wedge))>0$}
         \STATE $S'' \leftarrow {\tt apply}(S'',S^{\bm{\sigma}}_{X_1,\dots,X_\ell},\vee)$
            \STATE revised $\leftarrow$ true
         \ENDIF
    \ENDFOR
    \ENDFOR
\IF{{\bf not} revised}
\STATE  $\ell \leftarrow \ell+1$
\ENDIF
\ENDWHILE
\STATE $S''\leftarrow {\tt apply}(S'', S', \wedge )$
\STATE \textbf{return} $S''$
\end{algorithmic}
\end{algorithm}


\begin{theorem}\label{thm::bresdd}
Consider a KB $\psi$ over variables $\bm{X}$  and a new piece of information $\mu$ whose variables are in $\bm{X}$. Let $v$ be a vtree for $\bm{X}$, and $S$, $S'$ two SDDs normalized for $v$ representing $\psi$ and $\mu$, respectively. Algorithm \texttt{BRevSDD} returns a SDD representing  $\psi \circ \mu$. 
\end{theorem}

As expected, such general case has exponential worst-case complexity, as stated by the following result.

\begin{prp}\label{prop::complexitybrevdd}
\texttt{BRevSDD} runs in time $O(|S|^{4^n} + 4^n |S||S'|)$.
\end{prp}

A simple modification of Algorithm \ref{alg:BRevSDD} might prevent such a worst-case exponential growth. Let k be the order of the revision. Instead of building a single SDD by the disjunction of the compatible $k$-semi-resolvents (line 8), we might cope with a collection of SDDs, one for each compatible $k$-semi-resolvent and corresponding to the conjunction between the $k$-semi-resolvent and $S'$. The revised model should  therefore be intended as the disjunction of these $\frac{n!}{k!(n-k)!}2^k $ models, which is $O((2n)^k)$. Compatibility with a formula can be addressed by checking the compatibility with each element of such collection in $O((2n)^k|S||S'|)$.

In practical revision tasks, we might set a 
bound $\hat{k} \ll n$ for the maximal relaxation level applied to $\psi$. This corresponds to actually perform Dalal revision only if this can be achieved at order $k \leq \hat{k}$. If this is not the case, we consider the incompatibility between the $\mu$ and $\psi$ too high and go for an expansion (or a contraction).

Faster revision schemes can be achieved for specific classes of formulae. Algorithm \ref{alg:BRevSDDDNF2} is a variant of Algorithm \ref{alg:BRevSDD} to be used when the new information $\mu$ is provided in
\emph{disjunctive normal form} (DNF, that is a disjunction of conjunctions of literals), while $\psi$ remains a SDD. By separately checking the compatibility between each clause $c$ of $\mu$ and the $k$-semi-resolvents of $\psi$, we obtain $\psi \circ \mu$ by simply disjoining a subset of the original clauses of $\mu$. 
This is formalized by the next Theorem \ref{thm::bresdddnf}.


\begin{algorithm}[htb!]
\caption{\texttt{BRevSDD-DNF} (Belief Revision in SDDs)}
\label{alg:BRevSDDDNF2}
\textbf{Input}: SDD $S$ on $\bm{X}$ norm. for $v$ and such that $\langle S \rangle = \psi$, SDDs $\{S_c\}_{c\in C}$ on $\bm{X}$ norm. for $v$ representing the disjunctive clauses $\{c\}_{c\in C}$ of $\mu$, a complete DNF.\\
\textbf{Output}: SDD $S''$ on $\bm{X}$ norm. for $v$ s.t. $\langle S'' \rangle = \psi \circ \mu$.\\
\begin{algorithmic}[1] 
\STATE $S'' \leftarrow {\tt compile}(\bot,v)$
\STATE revised $\leftarrow$ false
\STATE $\ell \leftarrow 1$
\WHILE{{\bf not} revised}
 \FOR{$\{X_1,\dots,X_\ell\} \subseteq \bm{X} $}
    \FOR{$\bm{\sigma} \in \{+,-\}^\ell$}
     \FOR{$c\in C$}
        \IF{${\tt satisfies}(S^{\bm{\sigma}}_{X_1,\dots,X_\ell},c)$}
         \STATE $S'' \leftarrow {\tt apply}(S'',S_c,\vee)$
            \STATE revised $\leftarrow$ true
         \ENDIF
         \ENDFOR
    \ENDFOR
    \ENDFOR
\IF{{\bf not} revised}
\STATE  $\ell \leftarrow \ell+1$
\ENDIF
\ENDWHILE
\RETURN $S''$
\end{algorithmic}
\end{algorithm}


\begin{theorem}\label{thm::bresdddnf}
Let $\psi$ be a $\mathcal{L}$-formula, $\bm{X}$ be the variables occurring in $\psi$, $v$ be a vtree for $\bm{X}$, and $S$ a SDD normalized for $v$ representing $\psi$. Consider a complete DNF $\mu=\mathop{\bigvee}_{c\in C} c$, whose variables are in $\bm{X}$, and $\{S_c\}_{c\in C}$ be SDDs normalized for $v$ representing its clauses. Algorithm \texttt{BRevSDD-DNF} returns a SDD representing $\psi \circ \mu$. 
\end{theorem}

As the clause satisfaction test only takes linear time with respect to the size of the SDD, it is a trivial exercise to show the following results.

\begin{prp}\label{prop::complexitybrevdddnf}
\texttt{BRevSDD-DNF} runs in time $O(4^n |S| + \mathop{\Pi}_{c\in C}|S_c| )$.
\end{prp}

Even in this case, if we do not force the algorithm to return a single SDD, we might simply modify line 11 in order to keep count of the clauses consistent with at least one $k$-semi-resolvent. In this way the algorithm returns a collection of clauses of a DNF equivalent to $\psi \circ \mu$ in time $O(4^n|S|)$.


The following example shows an application of such DNF-based revision.

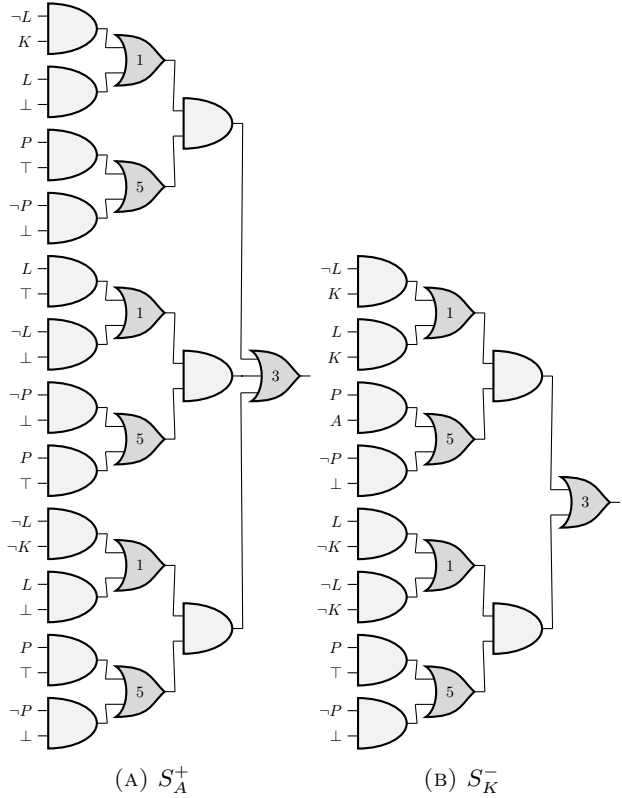
\begin{figure}[htp!]
\centering
\begin{subfigure}[b]{4cm}
\begin{tikzpicture}[scale=0.6,transform shape]
\draw (0,0) node[and port,fill=black!5] (myand1){};
\draw (0,-1.4) node[and port,fill=black!5] (myand2){};
\draw (0,-2.8) node[and port,fill=black!5] (myand3){};
\draw (0,-4.2) node[and port,fill=black!5] (myand4){};
\draw (0,-5.6) node[and port,fill=black!5] (myand5){};
\draw (0,-7.0) node[and port,fill=black!5] (myand6){};
\draw (0,-8.4) node[and port,fill=black!5] (myand7){};
\draw (0,-9.8) node[and port,fill=black!5] (myand8){};
\draw (0,-11.2) node[and port,fill=black!5] (myand9){};
\draw (0,-12.6) node[and port,fill=black!5] (myand10){};
\draw (0,-14) node[and port,fill=black!5] (myand11){};
\draw (0,-15.4) node[and port,fill=black!5] (myand12){};
\draw(1.5,-0.7) node[or port,fill=black!15] (myor1){1};
\draw(1.5,-3.5) node[or port,fill=black!15] (myor2){5};
\draw(1.5,-6.3) node[or port,fill=black!15] (myor3){1};
\draw(1.5,-9.1) node[or port,fill=black!15] (myor4){5};
\draw(1.5,-11.9) node[or port,fill=black!15] (myor5){1};
\draw(1.5,-14.7) node[or port,fill=black!15] (myor6){5};
\draw(3,-2.1) node[and port,fill=black!5] (myand1b){};
\draw(3,-7.7) node[and port,fill=black!5] (myand2b){};
\draw(3,-13.3) node[and port,fill=black!5] (myand3b){};
\draw(4.5,-7.7) node[or port,fill=black!15,number inputs=3] (root){3};
\draw (myand1.in 1) node[anchor=east] {$\neg L$};
\draw (myand1.in 2) node[anchor=east] {$K$};
\draw (myand2.in 1) node[anchor=east] {$L$};
\draw (myand2.in 2) node[anchor=east] {$\bot$};
\draw (myand3.in 1) node[anchor=east] {$P$};
\draw (myand3.in 2) node[anchor=east] {$\top$};
\draw (myand4.in 1) node[anchor=east] {$\neg P$};
\draw (myand4.in 2) node[anchor=east] {$\bot$};
\draw (myand5.in 1) node[anchor=east] {$L$};
\draw (myand5.in 2) node[anchor=east] {$\top$};
\draw (myand6.in 1) node[anchor=east] {$\neg L$};
\draw (myand6.in 2) node[anchor=east] {$\bot$};
\draw (myand7.in 1) node[anchor=east] {$\neg P$};
\draw (myand7.in 2) node[anchor=east] {$\bot$};
\draw (myand8.in 1) node[anchor=east] {$P$};
\draw (myand8.in 2) node[anchor=east] {$\top$};
\draw (myand9.in 1) node[anchor=east] {$\neg L$};
\draw (myand9.in 2) node[anchor=east] {$\neg K$};
\draw (myand10.in 1) node[anchor=east] {$L$};
\draw (myand10.in 2) node[anchor=east] {$\bot$};
\draw (myand11.in 1) node[anchor=east] {$P$};
\draw (myand11.in 2) node[anchor=east] {$\top$};
\draw (myand12.in 1) node[anchor=east] {$\neg P$};
\draw (myand12.in 2) node[anchor=east] {$\bot$};
\draw (myand1.out) -- (myor1.in 1) node[]{};
\draw (myand2.out) -- (myor1.in 2) node[]{};
\draw (myand3.out) -- (myor2.in 1) node[]{};
\draw (myand4.out) -- (myor2.in 2) node[]{};
\draw (myand5.out) -- (myor3.in 1) node[]{};
\draw (myand6.out) -- (myor3.in 2) node[]{};
\draw (myand7.out) -- (myor4.in 1) node[]{};
\draw (myand8.out) -- (myor4.in 2) node[]{};
\draw (myand9.out) -- (myor5.in 1) node[]{};
\draw (myand10.out) -- (myor5.in 2) node[]{};
\draw (myand11.out) -- (myor6.in 1) node[]{};
\draw (myand12.out) -- (myor6.in 2) node[]{};
\draw (myor1.out) -- (myand1b.in 1) node[]{};
\draw (myor2.out) -- (myand1b.in 2) node[]{};
\draw (myor3.out) -- (myand2b.in 1) node[]{};
\draw (myor4.out) -- (myand2b.in 2) node[]{};
\draw (myor5.out) -- (myand3b.in 1) node[]{};
\draw (myor6.out) -- (myand3b.in 2) node[]{};
\draw (myand1b.out) -- (root.in 1) node[]{};
\draw (myand2b.out) -- (root.in 2) node[]{};
\draw (myand3b.out) -- (root.in 3) node[]{};
\end{tikzpicture}
\caption{$S^+_A$\label{fig:semi1}}
\end{subfigure}
\begin{subfigure}[b]{4cm}
\begin{tikzpicture}[scale=0.6,transform shape]
\draw (0,0) node[and port,fill=black!5] (myand1){};
\draw (0,-1.4) node[and port,fill=black!5] (myand2){};
\draw (0,-2.8) node[and port,fill=black!5] (myand3){};
\draw (0,-4.2) node[and port,fill=black!5] (myand4){};
\draw (0,-5.6) node[and port,fill=black!5] (myand5){};
\draw (0,-7.0) node[and port,fill=black!5] (myand6){};
\draw (0,-8.4) node[and port,fill=black!5] (myand7){};
\draw (0,-9.8) node[and port,fill=black!5] (myand8){};
\draw(1.5,-0.7) node[or port,fill=black!15] (myor1){1};
\draw(1.5,-3.5) node[or port,fill=black!15] (myor2){5};
\draw(1.5,-6.3) node[or port,fill=black!15] (myor3){1};
\draw(1.5,-9.1) node[or port,fill=black!15] (myor4){5};
\draw(3,-2.1) node[and port,fill=black!5] (myand1b){};
\draw(3,-7.7) node[and port,fill=black!5] (myand2b){};
\draw(4.5,-4.9) node[or port,fill=black!15] (root){3};
\draw (myand1.in 1) node[anchor=east] {$\neg L$};
\draw (myand1.in 2) node[anchor=east] {$K$};
\draw (myand2.in 1) node[anchor=east] {$L$};
\draw (myand2.in 2) node[anchor=east] {$K$};
\draw (myand3.in 1) node[anchor=east] {$P$};
\draw (myand3.in 2) node[anchor=east] {$A$};
\draw (myand4.in 1) node[anchor=east] {$\neg P$};
\draw (myand4.in 2) node[anchor=east] {$\bot$};
\draw (myand5.in 1) node[anchor=east] {$L$};
\draw (myand5.in 2) node[anchor=east] {$\neg K$};
\draw (myand6.in 1) node[anchor=east] {$\neg L$};
\draw (myand6.in 2) node[anchor=east] {$\neg K$};
\draw (myand7.in 1) node[anchor=east] {$P$};
\draw (myand7.in 2) node[anchor=east] {$\top$};
\draw (myand8.in 1) node[anchor=east] {$\neg P$};
\draw (myand8.in 2) node[anchor=east] {$\bot$};
\draw (myand1.out) -- (myor1.in 1) node[]{};
\draw (myand2.out) -- (myor1.in 2) node[]{};
\draw (myand3.out) -- (myor2.in 1) node[]{};
\draw (myand4.out) -- (myor2.in 2) node[]{};
\draw (myand5.out) -- (myor3.in 1) node[]{};
\draw (myand6.out) -- (myor3.in 2) node[]{};
\draw (myand7.out) -- (myor4.in 1) node[]{};
\draw (myand8.out) -- (myor4.in 2) node[]{};
\draw (myor1.out) -- (myand1b.in 1) node[]{};
\draw (myor2.out) -- (myand1b.in 2) node[]{};
\draw (myor3.out) -- (myand2b.in 1) node[]{};
\draw (myor4.out) -- (myand2b.in 2) node[]{};
\draw (myand1b.out) -- (root.in 1) node[]{};
\draw (myand2b.out) -- (root.in 2) node[]{};
\end{tikzpicture}
\caption{$S^-_{K}$\label{fig:semi2}}
\end{subfigure}
\caption{Two SDD semiresolvents of $\psi$ in Equation (\ref{eq:kbkisa}).}
\label{fig:toy_semi}
\end{figure}

\begin{example}
Consider the SDD $S$ in Example \ref{ex:kisa} corresponding to the KB in Equation (\ref{eq:kbkisa}) and depicted in Figure \ref{fig:toy}. As a new piece of information, let us take $\mu := (\neg L \wedge K \wedge \neg P \wedge \neg A )\vee (L\wedge \neg K \wedge \neg P \wedge A)$. In order to obtain $\psi \circ \mu$,
we first apply Algorithm \ref{alg:ReSDD} for each $X\in \{L,K,P,A\}$. Figure \ref{fig:toy_semi} depicts for instance two semi-resolvents obtained in this way. Note that, by construction, these are SDDs normalized for the same vtree as the original SDD in Figure \ref{fig:toy}. Finally, when applying Algorithm \ref{alg:BRevSDDDNF2} we eventually obtain $\psi \circ \mu = \mu $ itself. Indeed both clauses of $\mu$ are compatible with at least one semi-resolvent: the first clause is compatible with $\psi_L^+$ and $\psi_P^-$, the second with $\psi_P^+$ and $\psi_A^-$. One can actually show that, for this particular $\psi$, any conjunction of four literals for $L,K,P,A$ inconsistent with $\psi$ is consistent with at least one of the semi-resolvents of $\psi$.
\end{example}


Let us conclude this section with an additional note on complexity. Exponential running time as a worst case seems to be unavoidable, given the repeated applications of {\tt apply} in \texttt{BRevSDD} and of $\texttt{SReSDD}^{\pm}$ in the general case. Notice that this happens for BDDs too. In \cite{gorogiannis2002implementation}, Gorogiannis and Ryan argued that empirical evidence indicates that the practical efficiency of revision with BDDs is much better than their worst case  $O(2^{6n}|S|)$.

\section{Experiments}
For a preliminary validation of the belief revision algorithm proposed in the previous section, we consider a synthetic benchmark based on randomly generated \emph{conjunctive normal forms} (CNFs), i.e., conjunctions of disjunctions of randomly picked literals. We cope with a set $\bm{X}$ of $n$ Boolean variables. As input KB $\psi$ we consider a CNF obtained by the conjunction of $n/2$ clauses, each involving three random literals. The same setup is used to generate the new formula $\mu$. A CNF over $n$ variables and treewidth $tw$ can be canonically represented by a SDD of size $O(n2^{tw})$ \cite{darwiche2011sdd}. We do not estimate the CNF treewidth, while simply setting a timeout for the SDD compilation. We consider only non-trivial revision tasks such that $\mu \not \models \psi$ (as otherwise we simply have $\psi \circ \mu = \mu$). For easier comparisons of the results, we only consider revisions achieved after a single iteration, i.e., $\psi \circ \mu  = G^1(\psi)\wedge \mu$.

The CNF associated with $\psi$ is first represented as a SDD $S$ normalized for a \emph{balanced} vtree (i.e., the sizes of the left and right sub-trees differ by at most one and both subtrees are balanced). Algorithm \ref{alg:BRevSDD} is used to compute $\psi \circ \mu$. As a proxy indicator of the potential of our approach we consider the SDD size. Such value is compared with the size of the SDD obtained by compilation of the formula corresponding to the manual implementation of the revision.

\begin{figure}[htp!]
\centering
\begin{tikzpicture}[scale=0.8]
\begin{axis}[
legend pos=north west,
xlabel=n,ymode = log,ylabel=$|S|$]
\addplot[color=red,mark=*]
plot[error bars/.cd,y dir=both, y explicit]
coordinates {
(10,323.46)+-(0.0,117.68013657986019)
(12,601.26)+-(0.0,181.39468770411312)
(14,1058.79)+-(0.0,337.2800980571705)
(16,1709.79)+-(0.0,684.901619345303)
(18,2725.26)+-(0.0,1168.6417275079336)
(20,4660.77)+-(0.0,1829.3224326857398)
(22,7952.04)+-(0.0,3137.078689713496)
(24,13193.4)+-(0.0,6663.422479691434)
(26,20773.26)+-(0.0,11097.738899709366)
(28,33233.64)+-(0.0,15167.607492661082)
(30,42509.72727)+-(0.0,17934.16387)
};
\addplot[color=blue,mark=square] 
plot[error bars/.cd,y dir=both, y explicit]
coordinates {
(10,207.03)+-(114.85128328724618,114.85128328724618)
(12,334.38)+-(0.0,169.29770443592176)
(14,583.95)+-(0.0,289.01348663089317)
(16,997.32)+-(0.0,518.8710331969874)
(18,1309.41)+-(0.0,788.0155114993843)
(20,1696.47)+-(0.0,830.9799795203469)
(22,2422.8)+-(0.0,1088.5929199910063)
(24,3536.52)+-(0,1890.38493261806)
(26,5364.48)+-(0.0,3508.2762046571056)
(28,7330.74)+-(0.0,4245.878408719761)
(30,10262.27)+-(0.0,5088.871665)
};
\legend{Revision + Compilation,Compilation + Revision}
\end{axis}
\end{tikzpicture}
\caption{Revised SDD sizes.}
\label{fig:experiments}
\end{figure}
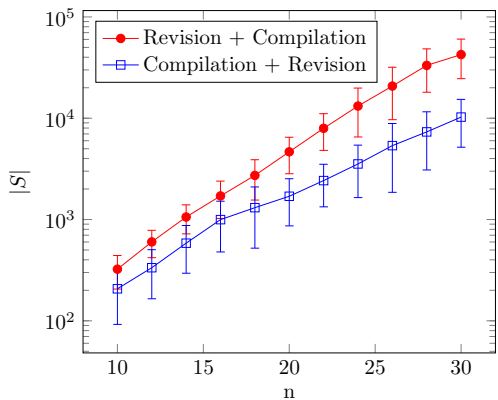

Experiments are performed within the \emph{Juice} library \cite{DangAAAI21}, a Julia tool for logic and probablistic circuits providing extensive SDD support. The Juice-based Julia implementation of our algorithms is freely available together with the code used for the simulations in Github (\href{https://github.com/IDSIA-papers/2021-KR-revision}{\tt IDSIA-papers/2022-BReSDD}). Figure \ref{fig:experiments} depicts the average circuit size for the two revisions approaches and increasing number of variables. Each point is an average over 100 random generations of the formulae and a timeout of one minute has been used. 
Despite the high variance due to the lack of a treewidth evaluation of the input formulae, the behaviour is clear and confirms our expectations: being based on local transformations, performing the revision in the SDD makes the revised SDD smaller than a compilation of the revised formula. Finally, let us note that in our experiments the sizes of the two SDDs remained unaffected by the choice of performing the intersection with $\mu$ as a last step of the revision process or right after the computation of each resolvent. A theoretical investigation of such empirical evidence is a necessary future work.

\section{Conclusions}
A very first belief revision scheme for SDDs has been presented. Contrary to the procedure adopted 
 by \cite{gorogiannis2002implementation} in case of BDDs, in this paper we exploited a syntactic characterisation of Dalal revision.  As a necessary future work we intend to refine the characterisation of our approach both at the theoretical level and by means of a dedicated empirical analysis. Another natural direction would be to adapt to SDDs the semantic approach employed in \cite{gorogiannis2002implementation}.
 A challenging  outlook concerns the application of this scheme to probabilistic sentential decision diagrams \cite{kisa2014probabilistic}, an important class of probabilistic circuits based on SDDs. 

\appendix
\section{Proofs}

\paragraph{Proof of Proposition \ref{lemma::resk}}
By definition $\psi^{\bm{\sigma}}_{X_1,\dots, X_{\ell}}$ has variables $X_1,\dots,X_{\ell}$ instantiated (according to $\bm{\sigma}$), thus it does not contain such variables. Moreover, assume that $w=\{X_1,\dots,X_m\}$ is a model of $\psi$. Then, if $m\geq \ell$, take any subset $\{X_{i_1},\dots,X_{i_{\ell}}\}$ of $w$. Let $\bm{\sigma}=(+,\dots,+) \in \{+,-\}^{\ell}$. Then we have that $w$ models

$$\psi^{\bm{\sigma}}_{X_{i_1},\dots,X_{i_{\ell}}}\wedge \mathop{\bigwedge}_{i_1\leq i \leq i_{\ell}}X^{\bm{\sigma}(i)}_{i}.$$

If, instead, $m<\ell$, take $\bm{\sigma} \in \{+,-\}^{\ell}$ such that $\bm{\sigma}(i)=+$ if $i\leq m$ and $\bm{\sigma}(i)=-$ otherwise. Then $w$ models

$$\psi^{\bm{\sigma}}_{X_1,\dots,X_{\ell}}\wedge \mathop{\bigwedge}_{1\leq i \leq \ell} X^{\bm{\sigma}(i)}_i.$$

On the other side, let $w$ be a model of the right-hand side of the equivalence. Then there exists variables $X_1,\dots,X_{\ell}\in \bm{X}$ and $\bm{\sigma}\in \{+,-\}^{\ell}$ such that 

$$w\models \psi^{\bm{\sigma}}_{X_1,\dots,X_{\ell}}\wedge \mathop{\bigwedge}_{1\leq i \leq \ell} X^{\bm{\sigma}(i)}_i.$$

This means that $w$ is a model of both $\psi^{\bm{\sigma}}_{X_1,\dots,X_{\ell}}$ and $\mathop{\bigwedge}_{1\leq i \leq \ell} X^{\bm{\sigma}(i)}_i$, which immediately yields that $w\models \psi$.
\qed 


\paragraph{Proof of Proposition \ref{prop::distributivity}.}
We first prove that $G^1(\psi_1 \vee \dots \vee \psi_m) = G^1(\psi_1)\vee \dots \vee G^1(\psi_m)$. In order to do so, we need to show that $mod(G^1(\psi_1)\vee \dots \vee G^1(\psi_m))= g^1(mod(\psi_1 \vee \dots \vee \psi_m))$. Now, let $w$ be a $\mathcal{L}$-interpretation. We have that $w$ is a model of $G^1(\psi_1)\vee \dots \vee G^1(\psi_m)$ iff $w\in mod(G^1(\psi_j))=g^1(mod(\psi_j))$ for some $1\leq j \leq m$. Since $g^1(mod(\psi_j))= \mathop{\bigcup}_{w'\in mod(\psi_j)}g^1(w')$, this is equivalent to say that $w\in g^1(w')$ for some $w'\in mod(\psi_j)$, for some $1\leq j \leq m $. 

Now, $mod(\psi_1 \vee \dots \vee \psi_m) = \mathop{\bigcup}_{j=1}^m mod(\psi_j)$, hence $g^1(mod(\psi_1 \vee \dots \vee \psi_m))=\mathop{\bigcup}_{w'\in \mathop{\bigcup}_{j=1}^m mod(\psi_j)}g^1(w').$ Thus, the base case is proved. Fix $\ell>1$ and assume the claim true for $j<\ell$. We have:
\begin{equation}
\begin{aligned}
&G^{\ell}(\psi_1 \vee \dots \vee \psi_m)\\
& = G^{\ell-1}(G^1(\psi_1 \vee \dots \vee \psi_m))\\
&= G^{\ell-1}(G^1(\psi_1)\vee \dots \vee G^1(\psi_m))\\
&= G^{\ell-1}(G^1(\psi_1))\vee \dots \vee G^{i-1}(G^1(\psi_m))\\
&= G^{\ell}(\psi_1)\vee \dots \vee G^i(\psi_m)\,. \qed
\end{aligned}
\end{equation}

\paragraph{Proof of Corollary \ref{cor::Gksemires}}

The proof is by induction on $\ell$. The base case, $\ell=1$, is true by Proposition \ref{thm::dalal}. Now, assume both the equivalences of the corollary true for $1<i<\ell$. Then $G^{\ell}(\psi)  =  G^1(G^{\ell-1}(\psi))$, which, by induction hypothesis is equivalent to
    $$G^1(\mathop{\bigvee}_{X_1,\dots,X_{\ell-1}\in \bm{X}}res_{X_1}(\dots (res_{X_{\ell-1}}(\psi))\dots)).$$
Thanks to the Proposition \ref{prop::distributivity}, the latter can be rewritten as
  $$\mathop{\bigvee}_{X_1,\dots,X_{\ell-1}\in \bm{X}} G^1(res_{X_1}(\dots (res_{X_{\ell-1}}(\psi))\dots)),$$
which, thanks to Proposition \ref{thm::dalal}, is equivalent to

$$ \mathop{\bigvee}_{X_1,\dots,X_{\ell-1}\in \bm{X}} \mathop{\bigvee}_{X_{\ell}\in \bm{X}}  res_{X_\ell}(res_{X_1}(\dots (res_{X_{\ell-1}}(\psi))\dots)).$$
Finally, thanks to Proposition \ref{prop::properties}, this rewrites as 
$$ \mathop{\bigvee}_{X_1,\dots,X_{\ell}\in \bm{X}}   res_{X_1}(\dots (res_{X_{\ell}}(\psi))\dots),$$
which proves the first equivalence. In order to see that the second equivalence holds too, it is enough to notice that $res_{X_1}(\dots (res_{X_{\ell}}(\psi))\dots)\equiv res_{X_1,\dots,X_\ell}$ (Equation \ref{eq::resk}), which is equal to $\mathop{\bigvee}_{\bm{\sigma}\in \{+,-\}^\ell}\psi^{\bm{\sigma}}_{X_1,\dots,X_\ell}$ by definition.
\qed

\paragraph{\bf Proof of Theorem \ref{thm::resdd}.} Thanks to the properties of {\tt apply}, 
it is enough to show that $S_X^+$ and $S_X^+$ are SDDs normalized for $v$ representing $\psi_X^+$ and $\psi_X^-$, respectively. Sub-SDDs $\{i_m\}_{m=1}^M$ (line 4) is the output of ${\tt decisions}(w,S)$, i.e., the decision nodes of $S$ normalized for $w$, that is the parent of $X$ in $v$ (line 3). We distinguish the two cases where $X$ is a left and a right child of $w$.
\begin{enumerate}[(1)]
\item If $X=w^l$ (line 6), $i_m = \{(X,s_m^+),(\neg X,s_m^- )\}$, for each $m=1,\ldots,M$, .
\item If $X=w^r$ (line 10), $i_m = \{(p^t_m,s^t_m)\}_{t=1}^{L(m)}$, for each $m=1,\ldots,M$, where $s^t_m \in \{X, \neg X, \top, \bot \}$ for each $t=1,\ldots,L(m)$.
\end{enumerate}
Let $\{j_k\}_{k=1}^K$ be the output of ${\tt decisions}(S,u)$ with $u$ parent of $w$ in $v$. Write $j_k=\{(p^l_k,s^l_k)\}_{l=1}^{L(k)}$, for each $k=1,\ldots,K$. We distinguish the two case where $w$ is a left and a right child of $u$.
\begin{enumerate}[(a)]
\item If $w = u^r$, the subs $\{s_k^l\}_{l=1}^{L(k)}$ of $j_k$ are sub-SDDs defined over the variables in $w$. Thus, for each $l=1,\ldots,L(k)$, we write $s^l_k = i_{m(k,l)}$.
\item If $w = u^l$  the primes $\{p_k^l\}_{l=1}^{L(k)}$ are sub-SDDs defined over the variables in $w$. Thus, for each $l=1,\ldots,L(k)$, we write $p^l_k=i_{m(k,l)}$.
\end{enumerate}
Overall we have four possible joint cases, to be denoted in the following as (1a), (1b), (2a) and (2b). In the procedure leading to $S^+$ or $S^-$ from $S$, we modify nodes $i_1, \ldots, i_M$ normalized for $w$ in $v$ (lines 7-8 or 13-15). As a consequence, the nodes $j_1, \ldots, j_K$ normalized for $u$ in $v$ are modified too. We denote the transformed nodes as $\{\hat{i}_m\}_{m=1}^M$ and $\{\hat{j}_k\}_{k=1}^K$, with $\hat{j}_k= \{ (\hat{p}^l_k, \hat{s}^l_k )\}_{l=1}^{L(k)}$ for each $k=1,\ldots,K$. It is straightforward to verify that each $\hat{i}_m$ is still a SDD normalized for $w$.

Let us first show that $S^+$ and $S^-$ are well defined SDDs. This corresponds to prove that: (i) $\hat{p}^l_k$ and $\hat{s}^l_k$ are SDDs over $u^l$ and $u^r$, respectively; and (ii) $\{\hat{p}^l_k\}_{l=1}^{L(k)}$ are a partition (i.e., they are mutually exclusive and exhaustive). Let us focus on the proofs for $S^+$, those for $S^-$ being analogous.
\begin{itemize}
\item \underline{Case (1a)}. As $X=w^l$ and $w=u^r$, for each $k=1,\ldots,K$, $j_k = \{ (p_k^l, i_{m(k,l)})\}_{l=1}^{L(k)}$ with $i_{m(k,l)} = \{ (X,s^+_{m(k,l)}),(\neg X, s^-_{m(k,l)})\}$. 
\begin{enumerate}[(i)]
\item We have $\hat{p}^l_k=p_l^k$ for each $l=1,\ldots,L(k)$. Being as in $S$, these primes are SDDs normalized for $u^l$. By construction, for each $l=1,\ldots,L(k)$, $\hat{s}^l_k = \hat{i}_{m(k,l)}$ is a SDD normalized for $u^r$, since we have only modified one of its subs by replacing it with a copy of the other one.
\item As the primes $\{\hat{p}^l_k\}_{l=1}^{L(k)}$ are those of $S$, they remain exclusive and exhaustive.
\end{enumerate}
\item \underline{Case (1b)}. As $X=w^l$ and $w=u^l$, for each $k=1,\ldots,K$, $j_k=\{( i_{m(k,l)}, s_k^l,)\}_{l=1}^{L(k)}$ with $i_{m(k,l)} = \{ (X, s^+_{m(k,l)}), (\neg X, s^-_{m(k,l)})\}$.  
\begin{enumerate}[(i)]
\item The proof is analogous to that of (i) for Case (1a), with the primes playing the role of the subs and vice versa.
\item Primes $\{{p}_k^l\}_{l=1}^{L(k)}$ are exclusive by definition. Thus, for each $l_1,l_2 = 1,\ldots,L(k)$ with $l_1 \neq l_2$:
\begin{equation}
\begin{aligned}
\left[(X \wedge \langle s^+_{m(k,l_1)}\rangle) \vee (\neg X\wedge \langle s^-_{m(k,l_1)}\rangle) \right] &\bigwedge \\\left[(X \wedge \langle s^+_{m(k,l_2)}\rangle) \vee (\neg X \wedge\langle s^-_{m(k,l_2)}\rangle) \right] &\equiv \bot\,,
\end{aligned}   
\end{equation}
and hence:
\begin{equation}
\begin{aligned}
\left[X \wedge \langle s^+_{m(k,l_1)}\rangle \wedge \langle s^+_{m(k,l_2)}\rangle\right] & \bigvee\\
\left[\neg X \wedge \langle s^-_{m(k,l_1)}\rangle \wedge \langle s^-_{m(k,l_2)}\rangle\right] &\equiv \bot\,.
\end{aligned}   
\end{equation}
This yields $\langle s^+_{m(k,l_1)}\rangle \wedge \langle s^+_{m(k,l_2)}\rangle = \bot$, this meaning that $\langle \hat{p}^{l_1}_k \rangle \wedge \langle \hat{p}^{l_2}_k \rangle \equiv \bot$. This proves exclusivity. For exhaustivity, we have:
\begin{equation}
\bigvee_{l=1}^{L(k)} \left[(X \wedge \langle s^+_{m(k,l)}\rangle ) \vee (\neg X \wedge \langle s^-_{m(k,l)}\rangle) \right] \equiv \top\,.
\end{equation}
The left-hand expression remains a tautology when we evaluate it in $X=\top$, which in such case becomes:
\begin{equation}
\bigvee_{l=1}^{L(k)} \langle s^+_{m(k,l)}\rangle = \bigvee_{l=1}^{L(k)} \langle \hat{p}^l_k \rangle\,.
\end{equation}
\end{enumerate}
\item \underline{Case (2a)}. As $X=w^r$ and $w=u^r$, we have $j_k = \{ (p_k^l, i_{m(k,l)})\}_{l=1}^{L(k)}$ for each $k=1,\ldots,K$, and $i_{m(k,l)}=\{ (p^t_{m(k,l)}, s^t_{m(k,l)})\}_{t=1}^{L(m(k,l))}$ with $s^t_{m(k,l)} \in \{ X,\neg X,\top,\bot\}$.
\begin{enumerate}[(i)]
\item Primes $\{ \hat{p}^l_k\}_{l=1}^{L(k)}$ are as those in $S$, i.e., they are still SDDs normalized for $u^l$. By construction, the subs $\{\hat{s}^l_k\}_{l=1}^{L(k)}$ are the result of an instantiation on their subs, i.e., terminal SDDs. Thus, they remain SDDs normalized for $u^r$.
\item The proof is as in case (1a).
\end{enumerate}
\item \underline{Case (2b)}. As $X=w^r$ and $w=u^l$, for each $k=1,\ldots,K$, $j_k = \{ ( i_{m(k,l)}, s_k^l)\}_{l=1}^{L(k)}$ with $i_{m(k,l)} = \{ (p^t_{m(k,l)}, s^t_{m(k,l)})\}_{t=1}^{L(m(k,l))}$ and $s^t_{m(k,l)} \in \{X,\neg X,\top,\bot\}$.
\begin{enumerate}[(i)]
\item By construction, primes $\{\hat{p}^l_k\}_{l=1}^{L(k)}$ are obtained by instantiating their subs, i.e., terminal SDDs. Thus, they remain SDDs normalized for $u^l$. The subs $\{\hat{s}^l_k\}$ are those of $S$, thus still SDDs normalized for $u^r$. 
\item Primes $\{\hat{p}^l_k\}_{l=1}^{L(k)}$ are exclusive, thus, for each $l_1,l_2=1,\ldots,L(k)$, $l_1 \neq l_2$, by exclusivity of the primes $\{p^l_k\}_{l=1}^{L(k)}$ we have:
\begin{equation}
\begin{aligned}
\left[ \bigvee_{t_1=1}^{L(m(k,l_1))} (\langle p^{t_1}_{m(k,l_1)} \rangle \wedge \langle s^{t_1}_{m(k,l_1)} \rangle  ) \right] &\bigwedge\\
\left[ \bigvee_{t_2=1}^{L(m(k,l_2))} (\langle p^{t_2}_{m(k,l_2)} \rangle \wedge \langle s^{t_2}_{m(k,l_2)} \rangle  ) \right]  &\equiv \bot\,.
\end{aligned}
\end{equation}
Hence, no matter how we instantiate the $s^{t_1}_{m(k,l_1)}$ and the $s^{t_2}_{m(k,l_2)}$, two distinct newly obtained primes remain incompatible.
 For exhaustivity, since 
\begin{equation}
\bigvee_{l=1}^{L(k)} \left[ \bigvee_{t=1}^{L(m(k,l))} ( \langle p^t_{m(k,l)} \rangle \wedge \langle s^t_{m(k,l)} \rangle) \right]\equiv \top\,,
\end{equation}
the left-hand expression remains a tautology when we evaluate it in $X=\top$, in which case it becomes exactly 
\begin{equation}
\bigvee_{l=1}^{L(k)} \langle \hat{p}^l_k \rangle \equiv \top\,.
\end{equation}
\end{enumerate}
\end{itemize}
Thus, $S_X^+$ and $S_X^-$ are well-defined SDDs normalized for the same vtree $v$ of $S$. It remains to show that they represent indeed the semi-resolvents of $\psi$ for $X$, i.e., $\langle S^+_X \rangle \equiv \psi_X^+ $ and $\langle S^-_X \rangle \equiv \psi_X^- $. As a consequence of Corollary ref{cor::res}, we know that this is equivalent to show that $\langle S^+_X \rangle \equiv \psi^+_X = \langle S \rangle ^+_X$ and $\langle S^-_X \rangle \equiv \psi^-_X = \langle S \rangle ^-_X$. Following Definition \ref{def:sdd}, $\langle S \rangle$ is written as a recursive decomposition based on the vtree $v$ involving formulae $\{\langle i_m \rangle\}_{m=1}^M$, these being exactly the sub-formulae of $\psi$ involving $X$.  Thus, for instance, $\langle S \rangle ^+_X$ is obtained by replacing $\langle i_m \rangle$ with $\langle i_m \rangle^+_X$ for each $m=1,\ldots,M$. Hence, this reduces to show that $\langle i_m \rangle^+_X = \langle i'_m \rangle$ for $S_X^+$ and that $\langle i_m \rangle^-_X = \langle i'_m \rangle$ for $S_X^-$.  For each $m=1,\ldots,M$, if $X=w^l$, by definition $\langle i_m \rangle^+_X = s_m^+$ and by construction $\langle \hat{i}_m \rangle = (X \wedge \langle s_m^+ \rangle) \vee (\neg X \wedge \langle s_m^+ \rangle) =  \langle s_m^+ \rangle$;  if $X=w^r$, it is immediate since $\langle \hat{i}_m \rangle$ is obtained from $\langle i_m \rangle$ by instantiating $X=\top$. We analogously proceed for $S_X^-$.
$\qed$

\paragraph{Proof of Proposition \ref{prop::complexityresdd}.} 
If $X=w^l$, the $M$ nodes of $S$ normalized for $w$ are replaced by sub-SDDs already available in the input SDD $S$ (line 7). At line 8 a compression is made on the currently processed node. Thus, the running time will be less or equal to $|S|$. If $X=w^r$, for each of the $M$ nodes $i_1,\dots, i_M$ normalized for $w$, the non-trivial literals for $X$ are replaced by a constant. Again, a compression is made on the currently processed node. Thus the worst-case running time of the algorithm is $O(|S|)$.    $\qed$

\paragraph{Proof of Theorem \ref{thm::bresdd}.}
Let $k$ be the order of the revision (notice that such $k$ exists, the worst case being the one with $k=n$ and thus with $\psi \circ \mu \equiv \mu$). Algorithm \ref{alg:BRevSDD} calls Algorithm \ref{alg:ReSDD} in order to build the SDDs representing all the $k$-semi-resolvents of $\psi$. At lines 7-8, the $k$-semi-resolvents consistent with $\mu$ are disjoint one at a time via \texttt{apply}. The obtained SDD is then conjoint with $S'$ to get the output SDD $S''$. The correctness of the algorithm ($\langle S'' \rangle = \psi \circ \mu$) is an immediate consequence of Corollary \ref{cor::Gksemires} and the fact that, by definition, $\psi \circ \mu = G^k(\psi)\wedge \mu$.$\qed$


\paragraph{Proof of Proposition \ref{prop::complexitybrevdd}}
If $k\leq n$ is the order on which the revision is achieved, Algorithm \ref{alg:BRevSDD} performs an \texttt{apply} of all the $k$-semi-resolvents consistent with $\mu$. The consistency check is made on each conjunction between a $k$-semi-resolvent and $S'$. The number of these conjunctions is equal to the number of $k$-semi-resolvents, i.e., $\frac{n!}{k!(n-k)!}2^k$, and each of them costs $O(|S||S'|)$. Since the binomial coefficient is bounded by $2^n$ and $k$ is bounded by $n$, we get a cost of $O(4^n |S||S'|)$. 

As mentioned above, the number of $k$-semi-resolvents consistent with $\mu$ is $O(4^n)$, and each of them has size smaller or equal to $|S|$. Thus, the cost of such \texttt{apply} is $O(|S|^{4^n})$. $\qed$

\paragraph{Proof of Theorem \ref{thm::bresdddnf}.}
We have that $\psi\circ \mu = G^k(\psi)\wedge \mu$ is equivalent to

$$\left(\bigvee_{\substack{X_1,\dots,X_k\in \bm{X},\\ \bm{\sigma} \in \{+,-\}^k}} \langle S_{X_1,\dots,X_k}^{\bm{\sigma}} \rangle \right)\wedge \mathop{\bigvee}_{c\in \mathcal{C}} c $$ 
which, in turns, rewrites as

$$\bigvee_{\substack{X_1,\dots,X_k\in \bm{X},\\ \bm{\sigma} \in \{+,-\}^k,\\ c\in \mathcal{C}}}(\langle S_{X_1,\dots,X_k}^{\bm{\sigma}}\rangle \wedge c)\,.$$

Assuming that each $c\in \mathcal{C}$ is complete, if one term $\langle S_{X_1,\dots,X_k}^{\bm{\sigma}}\rangle \wedge c$ in the above disjunction is satisfiable, then it is equivalent to $c$. Now, in Algorithm \ref{alg:BRevSDDDNF2}, the check at line 9 ensures that only the SDDs relative to such terms are considered in the output. $\qed$

\paragraph{Proof of Proposition \ref{prop::complexitybrevdddnf}}
Analogously to what we proved for Proposition \ref{prop::complexitybrevdd}, the cost of the satisfiability checks at line 9 is $O(4^n|S|)$. We then need to add the cost of the \texttt{apply} of all the $S_c$ consistent with at least one $k$-semi-resolvent. Notice that a $S_c$ might be consistent with several $k$-semi-resolvents. Algorithm \ref{alg:BRevSDDDNF2} implicitly keeps in memory if a $S_c$ has already been added, so that the operation at line 11 performs at most $|C|$ execution of \texttt{apply}. The cost of the latter is $\mathop{\Pi}_{c\in C}|S_c|$. $\qed$


\bibliographystyle{plain}
\bibliography{biblio}



\end{document}